# Application of the Affinity Propagation Clustering Technique to obtain traffic accident clusters at macro, meso, and micro levels

Fagner Sutel de Moura[a*], Christine Tessele Nodari[a]

[a]Industrial Engineering Graduate Program –Federal University of Rio Grande do Sul, Porto Alegre, Brazil

**Abstract**
Accident grouping is a crucial step in identifying accident-prone locations. Among the different accident grouping modes, clustering methods present excellent performance for discovering different distributions of accidents in space. This work introduces the Affinity Propagation Clustering (APC) approach for grouping traffic accidents based on criteria of similarity and dissimilarity between distributions of data points in space. The APC provides more realistic representations of the distribution of events from similarity matrices between instances. The results showed that when representative data samples obtain, the preference parameter of similarity provides the necessary performance to calibrate the model and generate clusters according to the desired characteristics. In addition, the study demonstrates that the preference parameter as a continuous parameter facilitates the calibration and control of the model's convergence, allowing the discovery of clustering patterns with less effort and greater control of the results.

1. Introduction

Road safety is one of the main activities related to traffic management. This activity presents considerable complexity in dealing with safety measures due to the diversity of contexts, types of occurrences, and outcomes associated with road traffic accidents (RTAs). There are two fronts of action in road safety, one aimed at preventing RTAs and the other dedicated to mitigating problems brought to light by the repeated occurrence of accidents. The first approach is preventive, and the second is reactive (Mansourkhaki et al., 2017).

Preventive actions seek to adopt the best practices for designing safe roads; reactive efforts aim to identify places with a high occurrence of RTAs to adopt mitigating measures. Reactive actions claim consistent methods capable of indicating when a location with a high occurrence of accidents characterizes as an area of significant accident concentration and when such occurrences are random outcomes (Washington et al., 2018).

Although publications indicate good practices and protocols for identifying and treating Accident-Prone Locations (APLs) with powerful application tools (AASHTO, 2010; MassDOT, 2020), they cannot consistently implement such recommendations often. In urban areas, the multiplicity of contexts and omitted variables limits the use of specific approaches, such as Safety Performance Functions (SPFs), requiring the adoption of alternative methods for network screening (Elvik, 1988; Moura et al., 2022; Washington et al., 2018).

The literature dealing with the identification of APL is not unanimous due to the diversity of methods applied and the different ways of grouping data such as road segmentation, moving window segment, intersections, clusters, and spatial cells (Elvik, 1988; Moura et al., 2022; Washington et al., 2018). Another factor of limited consensus concerns the diversity of APL ranking criteria, which makes identifying APL even more complex (GEURTS and WETS, 2007).

Although there are currently several approaches for identifying APLs, their application requires some care to avoid misidentifying APLs. Such errors can occur mainly in urban areas with dynamic and heterogeneous contexts. In urban areas, the variety of modes of travel, transport systems, traffic management models, and distribution of land use patterns entails a considerable complexity when choosing the method of grouping accidents, APL identifying, and prioritizing critical areas (Washington et al., 2018).

Although widely used, the segregation of the urban road network into homogeneous segments and fixed-radius intersections can fail to represent the spatial distribution of phenomena associated with RTAs. As urban areas present complex arrangements, this results in different traffic compositions that impact the occurrence profile of RTAs not adherent to equidimensional units such as sweep radii or sections of the same length.

In this way, accidents grouped in units of UAs such as road segments or intersections with pre-defined dimensions can ignore distributions of RTAs that extrapolate extensions of defined scan radii or segments. It is also possible that distributions of RTAs focus on a boundary region between sections or intersections, diluting their effects between different units of analysis (Moura et al., 2022; Washington et al., 2018).

When applying spatial analysis to identify APL, these methods analyze spatial areas with homogeneous density or equidimensional spatial units. This approach analyzes units with a neighborhood relationship. When considering neighborhood densities and measures of the neighborhood association, spatial analyzes capture the distribution of associated RTAs, regardless of the dimension of the adopted spatial unit.

Spatial approaches have the advantage of evaluating locations of occurrence of RTAs as contiguous spaces characterized by neighborhood effects. The correlation between neighbors indicates such regions as conglomerates with homogeneous distributions of RTA occurrences or severity indices (Moura et al., 2022; Washington et al., 2018).

By eliminating the restriction of accident association to the topology of segments and intersections with pre-defined dimensions, UAs as clusters or cells provide a better characterization of locations with the occurrence of RTAs. This UAs, considering the pool strength of spatially associated phenomena allows clusters to represent areas of RTAs contained in segments, intersections, and transition areas, incorporating underlying arrangements capable of providing a more realistic representation of the area associated with RTAs (Aguero-Valverde and Jovanis, 2008; Cheng et al., 2018; Huang et al., 2017).

Clusters, when adopted, present distinct peculiarities of spatial cells. After identifying and classifying the clusters of RTAs based on their frequency or severity factors: the areas delimited by the cluster boundaries form polygons. These polygons describe the distribution of RTAs and allow for the intersection with risk factors spatially associated with the occurrence of RTAs, helping to identify the contribution of different risk factors (GEURTS and WETS, 2007).

As the clusters form areas defined from the occurrences of RTAs, the updating of accident data and the use of time windows allows adopt of before-and-after studies. Also, it allows identifying the spatial dynamics in terms of expansion and contraction of the areas of occurrence of RTAs and the temporal dynamics, as the variation in the density of occurrences and their severity index (GEURTS and WETS, 2007).

This work proposes the adoption of accident clustering using the Affinity Propagation Clustering method as an alternative to identify better representations where RTAs occur based on similarity distribution criteria (Frey and Dueck, 2007). When using clusters for this task, the model provides areas with boundaries that represent the areas with RTAs.

## 2. Literature Review

### 2.1 Background

Cluster analysis aims to partition data into classes to provide data groups that contribute to data reduction, hypothesis generation, hypothesis testing, and prediction based on group processes (Halkidi et al., 2001). Cluster analysis is the task of identifying significant patterns in high-dimensional data sets, even when their underlying composition is unknown (Moiane and Machado, 2018; Rodriguez et al., 2019). This process, which does not require a priori classification, identifies significant groups in large data sets through discriminant analysis that generates data partitions (Fraley and Raftery, 1998; Halkidi et al., 2001).

Although the distribution of clusters in the dataset can do ignored when starting the cluster analysis, each set is composed of different statistical distributions that give rise to the partitions. These distributions compose mixture models that define clusters based on similarities between elements and dissimilarities between partitions (Fraley and Raftery, 1998; Halkidi et al., 2001; Moiane and Machado, 2018).

Cluster analysis is a data partitioning process based on heuristic operations or statistical models. In this process, the choice of data partitioning methods is related to the problem of deciding the optimal number of clusters (Fraley and Raftery, 1998; Halkidi et al., 2001; Rodriguez et al., 2019).

Choosing appropriate features and a clustering algorithm is necessary to perform cluster analysis. The first one chooses the adequate features to encode the information that describes the nature of objects. The second consists of the most appropriate partition model for the characteristics of the features used (Halkidi et al., 2001).

These approaches consist of an learning process (Halkidi et al., 2001), being implemented through methods such; i) linkage, ii) allocation, iii) partition, iv) model-based, v) spectral, vi) density, and vii) subspace-based (Fraley and Raftery, 1998; Rodriguez et al., 2019). The different clustering methods are described below.

Also known as hierarchical methods, linkage algorithms are executed in stages, performing sequential partitioning observations. Through segregation or agglomeration processes, it forms dendrograms to partition and hierarchize data (Fraley and Raftery, 1998; Halkidi et al., 2001).

The allocation approaches perform exchange of objects between partitions at each iteration after observing the criteria of similarity between observations. As similarities are identified, allocation methods keep the number of partitions fixed during all iterations by redistributing objects between partitions. As these algorithms exchange objects between partitions, they require the final number of clusters in their initialization (Fraley and Raftery, 1998).

Partition approaches, based on criteria of distance and number of classes, aim to identify partitions based on the distance between objects and centroids (Rodriguez et al., 2019). Model-based algorithms run statistical frameworks that aim to optimize maximum likelihood. To do so, they assume that each data partition has multivariate normal distributions (Rodriguez et al., 2019).

Algorithms that adopt spectral approaches deal with dimensional spaces not adherent to linear discriminant models. They use adjacency structures through affinity matrices between objects to deal with this specificity (Rodriguez et al., 2019). Density algorithm approaches seek to evaluate neighborhood spaces, identifying density variations and spatial regularities of distributions (Rodriguez et al., 2019).

The last category of clustering algorithms is the subspace model. This approach, oriented to high-dimensional datasets, identifies similarities between objects using sample subsets to generate different data partitions (Rodriguez et al., 2019). Given the ability to identify categories of objects from unclassified data, the adoption of cluster analysis is present in several areas of research and business (Zhang et al., 2014)

## 2.2 Algorithms Requirements

Some clustering algorithms require the number of clusters to be discovered during the iteration process at initialization. Others demand initialization parameters such as the size of scan areas, minimum number of observations contained in each partition, number of correlated cells, or some homogeneity parameter (Costa et al., 2004; Halkidi et al., 2001).

As clustering algorithms require some initialization parameters, they become sensitive to such parameters. By using initialization parameters, clustering algorithms can converge to minimal locations when their parameters are not carefully selected (Costa et al., 2004).

Currently, cluster analysis requires algorithms capable of dealing with continuous data flows. This requirement requires algorithms to categorize new data according to pre-existing models while updating those models (Zhang et al., 2014). These new requirements call for more robust models with new tools to handle dynamic data (Rodriguez et al., 2019).

This work proposes using the Affinity Propagation Clustering (APC) algorithm to deal with the specificities that the identification of APLs requires and adopt a local discovery model with specific distributions of RTAs. According to related works, this algorithm deals well with the initialization parameters and presents an easy-to-execute calibration model.

## 2.3 Related Works

The APC was initially developed and tested in bioinformatics applications, which justifies its frequent application in the analysis of biological data. The original work on APC applied cluster analysis to identify faces in images and partition gene clusters from microarray data (Frey and Dueck, 2007). Another work applied a soft-constraint version of APC to analyze lymphoma and brain cancer data with excellent performance for noise reduction (Leone et al., 2007).

Also, on biological analyses, the implementation test of the APCluster package that implements the APC approach in R language performed the analysis of clusters of coiled coils sequences to identify clusters of dimeric and trimeric sequences, using quadratic similarity matrix. This work successfully identified cluster sequences demonstrating the feasibility of APC for exploratory sequence analysis (Bodenhofer et al., 2011).

Among the different applications of APC, initiatives have also emerged to identify patterns in streaming data. This approach allows monitoring changes in the structure of data distributions while updating the applied models (Zhang et al., 2014). As APC has shown promise on dynamic datasets of high dimensionality and less sensitivity to initialization parameters, studies from different areas of knowledge started to adopt APC in classification tasks.

Given the ability of APC to choose cluster instances, properly handle continuous data streams, and update models, many works have do devoted to applying APC to Vehicular Ad-doc Networks (VANETS) and Mobile Ad-Hoc Networks (MANET). In WI-FI network applications, the APC approach was applied in MANETs to dynamically deal with topology changes occurring in the creation of virtual backbones. The APC approach aimed to stabilize the election of the cluster head (CH) responsible for managing intra and inter-cluster communication controls, improving the energy efficiency of MANETs (Nabar and Kadambi, 2018).

Using simulated data to identify spectral clusters, the APC and its derivations were applied to test the influence of the vector of input preference parameters. The data showed significant efficiency in the speed to perform the tasks, mainly when the extension of the APC Partition Affinity Propagation is applied, which uses sub-blocks of matrices originating from the similarity matrix of the APC (Moiane and Machado, 2018).

One of the first traffic engineering studies to adopt APC applied APC to identify Level of Service (LOS) as an alternative to the Highway Capacity Manual (TRB, 2020). This study adopted an inventory of road data combined with speed and travel time data to generate a set of data to submit to the data partition process in different LOS through APC (Nguyen et al., 2016).

Another work used APC to detect anomalies in intelligent vehicle networks composing VANETs. This approach sought to improve the security of connected vehicle networks by electing a more reliable CH to manage clusters composed of vehicular platoons (Yang et al., 2016). Given the relevance of VANETs in the future development of mobility, a recent study discussed initiatives that apply APC in VANETs to obtain more excellent stability when solving clustering and CH selection tasks, minimizing execution time and errors (Talib, 2019).

As APC emulates message exchange between nodes in its clustering process, the similarity function of APC was reconstructed to introduce communication-related parameters so that only are elected as CH in VANETs, the vehicles with low relative mobility and good communication. This work presented a better stabilization performance of the clustering process than other tested algorithms (Bi et al., 2020).

With the growth of car-sharing systems, APC was used to logistics car rental fleets in large urban centers. The proposed model was applied to support long-term decisions regarding identifying optimal points for car-sharing depots. With this approach, locations were identified that minimize the walking distance of customers to the depots, enhancing real gains considering the investments made (Liu et al., 2018) .

With the advent of cross-industry of the automotive and telecommunications sectors focusing on automotive automation, new technological challenges have emerged focused on the convergence of advances on the internet of things (IoT), Big Data, and autonomous vehicles (AV). With the emergence of simulation-based communication systems works, integrating big data and IoT, the need arose to identify realistic styles based on driving styles of automotive controls. In this context, APC was used to group driving styles to be submitted to driving-data augmentation methods and then submit these results to convolutional neural network models to train new driving models (Zhang et al., 2019).

The emergence of V2X communication that includes vehicle-to-vehicle, vehicle-to-infrastructure, and vehicle-to-pedestrian connectivity presents new challenges. Such challenges consist in the feasibility of long- and short-range communication that operates interactively in different scenarios. In this context, APC proposes a dynamic model for identifying CH with the responsibility of acting as a proxy between vehicles and the public land mobile network in parallel with a machine-learning algorithm to manage the granularity of clusters dynamically (Koshimizu et al., 2020).

In addition to applications aimed at new technologies, APC does apply to improve pre-existing applications such as car-counting and classification. To improve the performance of this technology when embedded in low-performance computers. APC was applied to replace Nom-Maximum Suppression, a class of algorithms for selecting entities (bounding boxes) that represent vehicles on the road. In this scenario, APC was used to reduce redundant entities by grouping features from the combination of centroids (Harikrishnan et al., 2021).

3. **Affinity Propagation Clustering**

To apply a calibratable and update-oriented clustering model, the present work uses the APC approach. APC is an information extraction approach capable of generating partitions from similarities between pairs of data points containing both categorical and numerical data (Frey and Dueck, 2007; Moiane and Machado, 2018).

As reported in works that apply APC, the algorithm can apply approaches derived from different clustering algorithms. APC can handle with adjacency relationships through the use of similarity matrices. Furthermore, by using such matrices, APC can obtain sub-matrices that characterize sub-space models. Furthermore, the APC is update-oriented, which allows the implementation of models online. Finally, the APC, when dealing with granularities, allows dealing with density distributions of data points. Below are presented the specifics and the fundamental characteristics of the APC.

APC is a competitive learning algorithm that initializes each data point as a potential cluster exemplar. This strategy makes the centroid of each cluster, in fact, one of the constituent elements of the partition elected as cluster exemplar (Bi et al., 2020). Each exemplar acts as a cluster center that best describes the data partition (Shea et al., 2009).

For this purpose, APC uses a pairwise similarity function applied to data points, providing proximity criteria between objects contained in a similarity matrix (Nabar and Kadambi, 2018). This function seeks to minimize quadratic errors given by the distance between samples and data points of the same cluster (Hassanabadi et al., 2014). Once the similarity or relative attractiveness measures do calculate, the APC seeks to maximize the current evidence of each data point for how well-suited point k is to serve as the exemplar for i' elements of the cluster (Bhuyan and Mohapatra, 2014; Leone et al., 2007; Shea et al., 2009).

To carry out this process, the similarity matrix obtained $s(i, k)$ initially provides affinity values between data points. These affinities compose the content of messages exchanged recursively between data points. As in the first iteration, the availabilities are zero, and all data points are initialized as potential candidates, establishing an arrangement where all compete to be an exemplar.

For the message exchange process to create partitions that will give rise to clusters, the model uses a vector of input preferences to find preferable exemplars and cluster elements based on criteria of similarities and dissimilarities (Furtlehner et al., 2010). During this process, each point i sends the assigned responsibility r(i, k) to the others k' points, communicating the affinity or current evidence that i has that k' can act as its exemplar, considering the support of the others potential exemplars (Leone et al., 2007).

$$r(i,k) = s(i,k) - \max\{a(i,k') + s(i,k')\}$$

After, the k' rival points respond to their i emissary with availability a(i, k). The availability provides the accumulated evidence of how appropriate k' is to serve as an exemplar of i. For this, it considers the support offered by its self-responsibility r(k,k) and by the responsibility it presents to the other i' r(i',k) (Bhuyan and Mohapatra, 2014; Frey and Dueck, 2007). These iterations recursively update the similarity, responsibility, and availability matrices.

$$a(i,k) = min\left\{,0, r(k,k) + \sum \max\{0, r(i',k)\}\right\}$$

In addition to the similarity, availability, and responsibility between different data points, each object has its availability function initialized to zero. At the end of the iterations, those points affiliated with other exemplars remain available close to zero, reducing the evidence that they can be exemplars (Leone et al., 2007). This minimization of availabilities impacts the reduction of similarities s(i, k'), removing such data points from the competition for the exemplar position (Frey and Dueck, 2007; Leone et al., 2007).

$$a(k,k) = \sum \max\{0, r(i',k)\}$$

In addition to the self-availability function a(k,k), each data point has its self-responsibility r(k,k). Updating the self-availability and self-responsibility matrices when k=i indicates the accumulated evidence of each data point to act as an exemplar when the sum of both is positive (Hassanabadi et al., 2014; Shea et al., 2009):

$$CH_i = \arg\max\{a(i,k) + r(i,j)\}$$

This process ends when the cluster structure stabilizes by maximizing the similarities of all data points, and the decisions about exemplars become constant after a certain number of iterations (Frey and Dueck, 2007; Hassanabadi et al., 2014; Leone et al., 2007; Moiane and Machado, 2018; Shea et al., 2009). This process eliminates the need to define the number of clusters at process initialization. The algorithm finds the number of clusters from the similarity parameters contained in the vector of input preferences (Bodenhofer et al., 2011).

## 4. Methodology

The grouping of RTAs into UAs constitutes a crucial step for analyzing APLs. In this way, choosing an algorithm that corresponds to the need for data grouping and the formation of UAs determines the performance of APLs analysis.

To group RTAs as UAs, this work proposes the use of the competitive learning algorithm APC (BODENHOFER; KOTHMEIER; HOCHREITER, 2011; DUECK, 2009; FREY; DUECK, 2007). The use of the algorithm seeks to identify accident-prone locations in urban areas considering their spatial distribution pattern. The following is an overview of the work process using of APC summarized in Figure 1.

Figure 1. Steps of Work Process.

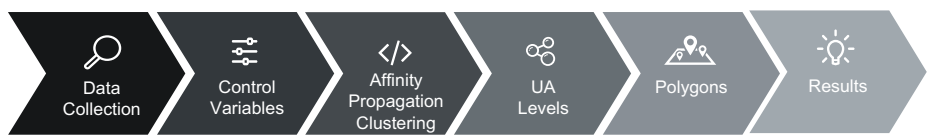

This work proposes the application of APC as a heuristic tool for data partitioning and UA obtaining by clustering through the variation of two control variables. For this purpose, was used the dataset corresponding to 6 years of RTAs records in Porto Alegre, available on the city's open data platform. Although the dataset has information on vehicle type, accident type, outcome, and severity, to carry out this work, the latitude-longitude coordinates of the occurrences were used to identify patterns of the spatial distribution of the RTAs.

This work evaluated the achievement of clusters from two control variables at different levels. The variables used are the sample size submitted to clustering (2 levels) and the similarity criterion (5 levels). The control variables correspond to the number of data points sampled from the complete data set, and the similarity criterion the exemplar preference parameter "q" provided to the vector of input preference for clustering process.

With different combinations of similarity criteria and sample sizes, we tested obtaining UAs of macro, meso, and micro levels at every combination of parameters (Barceló et al., 2005; Sanderson et al., 2012). As the definition of UA as meso and macro levels does not present consensus and has a qualitative component in its definition, this work suggests a delimitation parameter based on the number of intersections that constitute the UAs (Cai et al., 2019; Wang et al., 2021). The criterion for defining the different levels of UAs does obtain by the median number of intersections contained in the areas of the clusters.

The characterization of UAs at the micro-level considers spatial units where the RTAs relationships tend to be determined by individual actions or by groups of vehicles affected by local demand, conflicts, road geometry, or traffic control. When the combinations of parameters resulted in clusters with median intersections minor or equal to one, the classification of their UAs as micro-level was adopted (AASHTO, 2010; Cai et al., 2019; Wang et al., 2021).

As the data used originate from a city with 497 km$^2$, this work assumed that the meso level UAs have a dimension of 1 km$^2$. After, a grid with cells of 1 km$^2$ was generated, and the median of intersections contained in the generated cells was calculated. The median value was adopted as a limit to define meso level UAs. In the applied scenario, the median obtained corresponded to thirty intersections per km$^2$.

Meso-level UAs consist of locations influenced by the pool strength generated by the local road system, local economic profile, and associated traffic demand. When the combinations of parameters resulted in clusters with a median of intersections greater than one and less than or equal to thirty, the classification of their UAs as meso-level was adopted.

Macro-level UAs are spatial units that describe regional contexts influenced by road hierarchy, land-use profile, and regional insertion in the traffic system. As micro level UAs have the median of intersections less than or equal to one, and meso levels UAs have a median between two and thirty; when the combinations of parameters result in clusters with a median of intersections greater than thirty, their UAs classification is adopted as macro level.

After obtaining the clusters, they are treated by algorithms in the R programming language to transform the information into geographic polygon features. These features represent the spatial dimension that delimits the perimeter of the identified clusters. This approach allows measuring each polygon, its area, density, severity pattern, and other available spatial information.

For this activity, the contoreR library was used to obtain the perimeter points of each cluster and then define its contour (CRAN, 2015). After obtaining the contours of the clusters, the information of the data points affiliated to each cluster were persisted as SpatialPolygonsDataFrame using the SP library (Pebesm et al., 2021).

Combining data points and clusters in SpatialPolygonsDataFrame allowed treating cluster data as a Geographic Information System (GIS). GIS allows obtaining spatial visualizations and analysis of data. This result allows cluster data to be treated as spatial data, identifying their spatial association relationships. With this instrument, the area, and the median of intersections in the cluster do calculate.

## 4.1 Application to Data

This study proposed using the APC algorithm to apply a heuristic approach to identify accident distribution patterns using low-complexity feature. With the APC, the discriminant analysis process was performed using features that represent the latitude-longitude coordinates of the RTAs occurrences.

Each identified accident was submitted to a clustering process using the latitude-longitude features configured in the WGS 1984 coordinate system, EPSG:4326. The use of geographic coordinates abstracts the type and outcome of accidents and allows generating the identification of significant groups from the pattern of spatial distribution in terms of density and spatial proximity.

The use of geographic coordinates constitutes a set of features suitable for encoding information and identifying significant groups in a first-order analysis when considering spatial proximity and homogeneous density relationships. This first analysis makes it possible to choose groups for second-order verification, which consists of identifying the relationships underlying the data grouped in UAs (Moura et al., 2022).

The APC, when grouping accidents by criteria of spatial similarity and dissimilarity, provides identified clusters without the restriction of type, outcome, and network topology, allowing the identification of the pool strength of occurrences (Aguero-Valverde and Jovanis,

2008; Cheng et al., 2018; Huang et al., 2017; Lee et al., 2019). This approach generates uniform partitions in terms of distribution associated with local indicators that provide information to be understood as a system of underlying relationships.

After generating the clusters for the different levels of control variables, the number of clusters generated, the median of the spatial dimension in km$^2$, and the median of intersections per cluster were verified. This approach, by aggregating data in spatial units, allowed monitoring the algorithm's convergence in macro, meso, and micro level UAs, according to the criteria previously established.

As the heuristic approach used in this work does not have a validation criterion by previous classifications, the level scale of the UAS was determined by the median of intersections at clusters. The two sample levels of RTAs were submitted to the clustering process using five levels of the similarity of input preference parameter.

The similarity parameters of the vector of input preferences were 0.5, 0.75, 0.99, 0.999, and 1. The sample size was also applied at two levels, corresponding to 5000 and 10000 sample units from a universe of 70102 RTAs. The combination of two levels of sample size with five levels of similarity criterion resulted in 10 clustering parameter arrangements.

According to Moura et al. (2022), spatial cluster analyses consist of different micro, meso, and macro-level scales. Considering this specificity, this work sought to verify the number of clusters obtained, median area size, and median intersections per cluster among the ten verified arrangements. The median number of intersections per cluster was adopted to identify the feasibility of APC to control the different scales of clusters.

## 5. Results

As expected, the variation in the levels of preference (q) and the number of data points sampled affect the number and size of clusters obtained. The results obtained are presented below.

The number of data points used for clustering was kept at two levels for this experiment. Sampling levels were defined, containing 5000 and 1000 data points extracted from a set of 75102 RTAs.

The two sample sets of RTAs were subjected to clustering using the five levels of the similarity preference parameter. The results obtained were presented in a summary show in the Table 1 which demonstrates the relationship between the sample size and the input preference parameter on the configuration of the number of clusters obtained, the median of the cluster area and the median of intersections to each cluster.

As the number of intersections per UA was used to assess their scale, Table 2 shows the frequency of intersections per UAs obtained from different combinations of control variables. The results indicate that as the input preference parameter tends to 1, the configuration of the UAs assume micro level dimensions with up to one intersection per UA.

Finally, the Table 3 shows the spatial distribution of the identified clusters. Through the spatial representation it is also possible to perceive the contribution of the input preference parameter to define the scale of clusters, followed by the contribution of the sample size in this process.

As discussed in the methodological section, for the identification of clusters at macro, meso and micro-levels, the number of intersections per cluster was considered. Clusters with median up to 1 intersection were classified as micro level UAs, between 2 and 25 intersections were classified as meso level UAs, and UAs with more than 25 intersections were considered as macro-level UAs.

The similarity parameter represents the classification variable with the potential to calibrate the model and provide different cluster scales that result in the three levels of UAs. When using the parameters $q=0.5$ and $q=.75$ as a result, clusters emerged with a median number of intersections greater than 30 for the two adopted sample sizes, which generated macro-level UAs. When using the similarity parameter $q=0.99$ over the two sample sizes, the median number of intersections was between 2 and 30, characterizing meso-level UAs. Finally, when using the similarity parameters $q=0.999$ and $q=1$, clusters with a median of up to 1 intersection emerged, characterizing micro-level UAs as presented in Table 1.

The data demonstrate that the variation in the sample size has a smaller effect on the discovery different levels of clusters when the input preference parameter is less demanding. The number of clusters generated increased by doubling the sample size, generating elements with a smaller area and a smaller median of intersections per UA, without changing their scale when the input preference parameter is less demanding. The sample size only had an effect on the cluster scales when the input parameter tends to 1, this effect is perceived at input preference parameter levels with $q = (.999$ and $1)$.

At a level of $q=.5$ of the preference parameter, when doubling the sample size, the number of clusters went from 69 to 96. The median number of intersections reduced from 189 to 149, a variation of 39%, and -21%, respectively. When applying a $q=.75$ level when doubling the sample size, the number of clusters went from 104 to 150. The median number of intersections reduced from 127 to 77, a variation of 44% and -39%, respectively. For the two levels of the input reference parameter and sample size, the median of intersections per UA was greater than thirty, generating macro level UAs.

With a p=.99 level, when doubling the sample size. the number of clusters it went from 579 to 777, and the median number of intersections reduced from 18 to 13, the variation of 34% and -27%, respectively. Combining the input reference parameter and sample size at the levels presented resulted in a median of intersections per UA in the range between two and thirty, generating meso level UAs.

Using the p=.999 level, when doubling the sample size, the number of clusters went from 1935 to 2684. The median number of intersections reduced from 4 to 1, a variation of 38% and -75% respectively. Finally, when applying a p=1 level when doubling the sample size, the number of clusters went from 4520 to 8613. The median number of intersections remained at 1 in both models, with only the number of clusters varying by 90%. When adopted p=999, the median of intersections resulted in a meso level UA for a sample size of 5000 and a micro level for a sample size of 10000. For p=1, both sample sizes resulted in a median equal to 1, which configures micro-level UAs for both cases.

The sample size starts to influence the cluster pattern more sensitively when the input preference parameter tends to its maximum limit q=1, demonstrated in Table 1. The results demonstrate that when the sample size is fixed, the preference parameter has a potentiated effect on the configuration of the clusters. As shown in Table 2, which presents the median of intersections by UA; the increase in the similarity criterion accelerates meso and micro-level clusters' obtaining when the preference parameter tends to 1.

Table 1. Clusters summary.

|   | Clusters (n) | | Area km² | | Intersections (n) | |
| --- | --- | --- | --- | --- | --- | --- |
| P | 5000 | 10000 | 5000 | 10000 | 5000 | 10000 |
| .5 | 69 | 96 | 2.854 | 2.013 | 189 | 149 |
| .75 | 104 | 150 | 1.773 | 1.180 | 127 | 77 |
| .99 | 579 | 777 | 0.173 | 0.109 | 18 | 13 |
| .999 | 1935 | 2684 | 0.019 | 0.002 | 4 | 1 |
| 1 | 4520 | 8613 | 0.002 | 0.001 | 1 | 1 |

The cluster distribution pattern showed that the variation of the similarity parameter affects the size of the clusters, being accelerated when the parameter tends to 1. The Table 3 presents the visual distribution of clusters and demonstrates the effects of the similarity parameter and sample size to obtain different scales of clusters.

Table 2. Number of intersections associated with different cluster scales obtained.

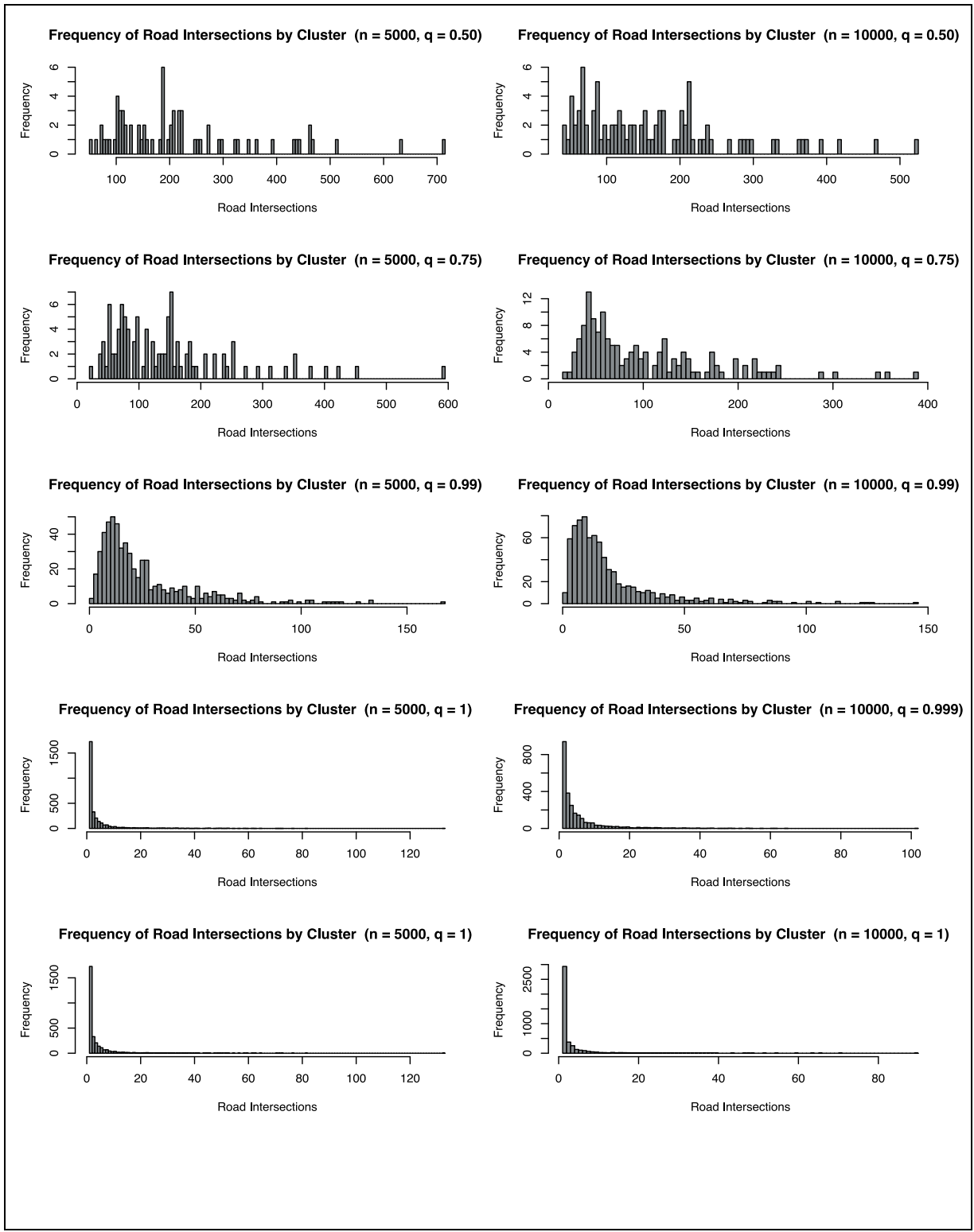

Table 3. Clusters emerged from different levels of cluster scales obtained.

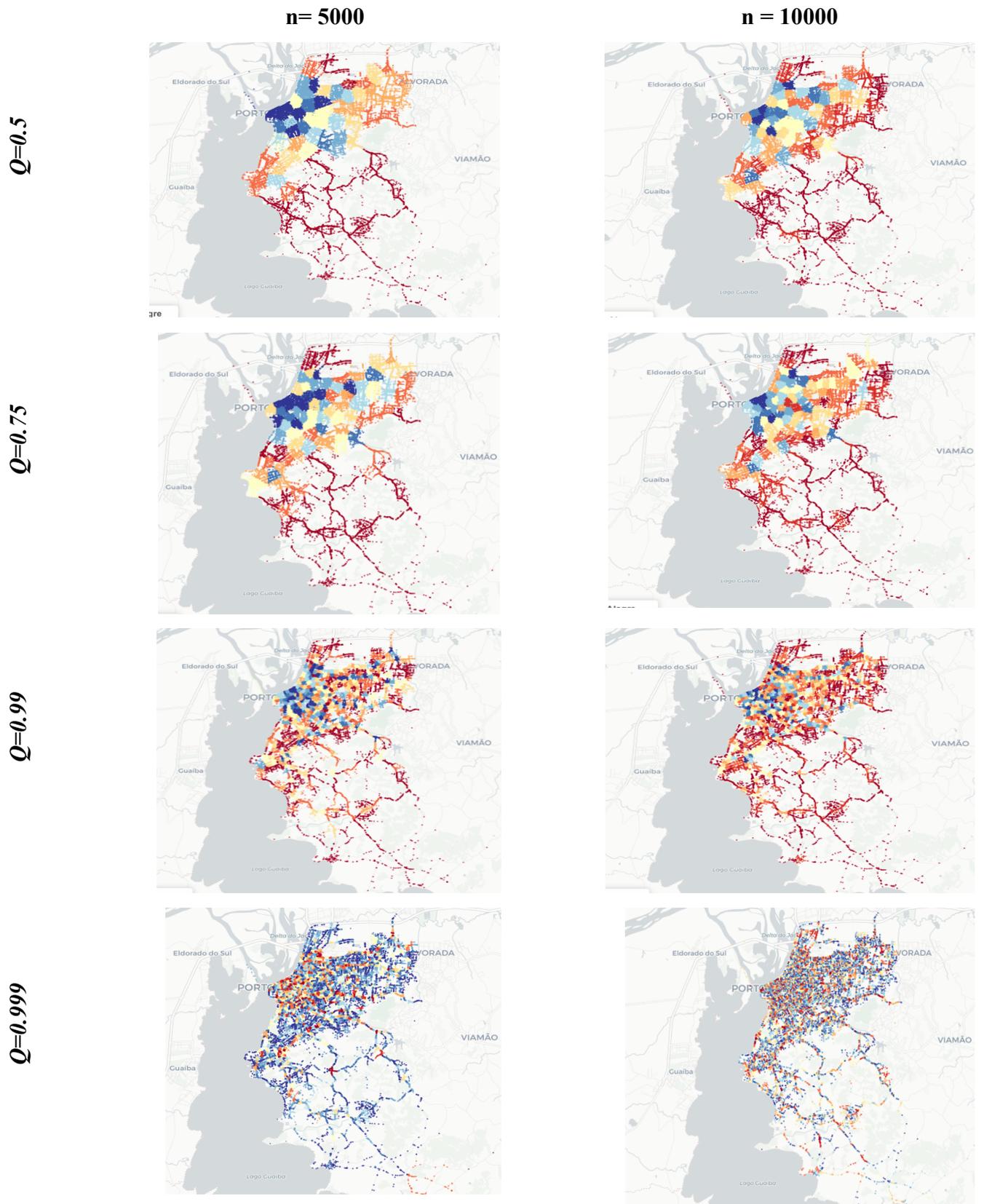

## 6. Discussion

As the data used for clustering do not have any previous classification regarding the definition of UAs from meso and macro level; this work proposed a scale to verify the performance of the APC. The ability of the algorithm to identify macro, meso, and micro-level UAs was verified by changing their preference parameter and the sample size submitted for clustering. The UAs levels were defined using the median of insertions contained in the spatial areas delimited by the obtained clusters.

When the combinations of parameters resulted in clusters with a median of intersections less than or equal to one, the classification of their UAs as micro-level was adopted. When the combinations of parameters resulted in clusters with a median of intersections greater than one and less than or equal to thirty, the classification of their UAs as meso-level was applied. When the combinations of parameters resulted in clusters with a median of intersections greater than thirty, the classification of their UAs was adopted as macro-level.

The search for different levels of UAs was performed by changing the levels of two control variables. The variables used were the preference parameter at five levels and the sample size at two levels. The value of the preference parameter varies in a continuum interval between zero and one; for the purposes of this work, are adopted the values 0.5, 0.75, 0.99, 0.999, and 1 to input preference parameter. The dataset used contains 75102 RTAs; two samples of sizes 5000 and 10000 were extracted to experiment. For each combination, the number of clusters generated, and the number of intersections contained in the areas covered by them were verified.

The results showed that when the input preference parameter value approaches 1, intra-cluster similarities and inter-cluster dissimilarities do maximize. With this behavior, the distribution of clusters is allocated, forming UAs of smaller spatial dimensions with a median of intersections less than or equal to one. As the size of the clusters reduces, they become part of the association of accidents at the local level, generating micro level UAs with high density and proximity between RTAs. This phenomenon was identified when the applied preference parameter was ($q=.999$) to sample size equal 10000 and ($q=1$) for both sample sizes.

When there is a relaxation in the preference parameter, the number of clusters reduces, and their area grows. This process delimits broader UAs, affiliating to each cluster RTAs distributed across multiple roads and intersections with a median of intersections greater than one and less than or equal to thirty. This arrangement encompasses accidents in areas with less connectivity to each other. They are characterizing regions of greater heterogeneity containing

RTAs in different classes of roads. This transition arrangement provides meso level UAs characterized by a hybrid spatial composition. This process occurred when the preference parameter applied was (q=.99) for both sample sizes and (q=.999) for sample size equal 5000.

When the input value of the preference parameter is reduced more faster, the precision of intra-cluster similarities and inter-cluster dissimilarities are reduced. In this way, a smaller number of UAs with greater spatial dimensions is obtained. With the expansion of the size of the clusters, they begin to compose the association of accidents at the regional level, composing macro-level UAs with a lower density of RTAs and very heterogeneous areas with a median of intersections greater than thirty. This result was obtained when the preference parameter applied was (q=.75) and (q=.5) for both sample sizes.

The results obtained show the contribution of the preference parameter (q) in increasing the number of clusters and reducing the average of intersections and area of UAs. The use of different levels of the preference parameter (q) allowed to modulate the performance of the APC algorithm. By modulating the APC and controlling its results, it was possible to generate clusters at different scales of UAs such as macro, meso, and micro level.

When different sample sizes were adopted, the contribution of the sample size to the generation of a greater number of clusters with a smaller spatial dimension was identified. Although different sample sizes interfere with the number of clusters and cluster area, the sample size at different levels did not imply the change in the UAs' scale when the input preference parameter adopts more relaxed values. The sample size contributed more effectively to the definition of cluster scales when adopted the maximization of value of the input preference parameter. In this case, the number of samples contributed more effectively to discover patterns, demonstrated by getting micro level UA for (q=999) and sample size equal to 10000.

These results demonstrate that the preference parameter is characterized as a more refined way of modulating the number and dimensions of clusters. It is shown that regardless of the sample size, the preference parameter provides a sufficient measure to calibrate a clustering model with APC, when using a representative sample. The preference parameter adopts a continuous scale in the range [0, 1] that allows for fine-tuning the composition of clusters. Performance with this adjustment level would demand much more time and effort if the trial-and-error approach used different sample sizes to calibrate the model. In addition, by increasing the sample size, the clustering process demands greater computational effort, demanding more time and hardware resources.

## 7. Conclusion

Clustering algorithms identify categories of objects from unclassified data. The APC is an unsupervised clustering algorithm that searches for similarities between objects. Due to the characteristics of APC, it was applied in this work to identify hidden patterns in RTAs distributions.

The patterns pursued consist of the different configurations in the distribution of RTAs. Three ideal distribution patterns of RTAs were adopted. First, the pattern of highly homogeneous cluster sets that maximize intra-cluster similarity and inter-cluster dissimilarity was adopted to identify specific distributions in a large dataset, providing micro level UAs.

The cluster sets of relative homogeneity patterns were adopted to identify local distributions, providing UAs for meso level analysis. Finally, sets of clusters of lesser homogeneity containing different data distributions were adopted to identify regional data distributions, providing UAs for macro level analysis.

This work verified the influence of sample size and preference parameters on the profile of clusters generated using APC. The results demonstrate that although the sample size affects the result, it does not affect the UAs levels when there is a relaxation in the preference parameter.

The adjustment needed to find sets of clusters that make up UAs between the different levels was obtained just by changing the preference parameter. Thus, the APC delivered the desired performance by applying just the calibration of the preference parameter.

The APC showed good performance in obtaining clusters from the preference parameter settings more easily than varying the sample size. Thus, with a representative sample of accidents in urban areas with a high concentration of RTAs, the use of preference parameter of APC provides an instrument for partitioning accidents adherent to local distribution pattern.

While other clustering algorithms adopt some arbitrariness to define clusters, APC allows the identification of areas corresponding to spatial distribution patterns only by applying variations in their similarity parameter calibrate as a continuum measure. This work demonstrated that APC makes it possible to obtain clusters from the maximization of the similarity criterion, providing partitions that emerge from the characteristics of the data with less effort and greater practicality.